\documentclass{article}

\makeatletter
\providecommand{\@noticestring}{}
\makeatother

\usepackage[preprint]{neurips_2023}
\usepackage{caption}
\usepackage[utf8]{inputenc} 
\usepackage[T1]{fontenc}    
\usepackage{hyperref}       
\usepackage{url}            
\usepackage{booktabs}       
\usepackage{amsfonts}       
\usepackage{nicefrac}
\usepackage{amsmath}
\usepackage{microtype}      
\usepackage{xcolor}         
\usepackage{xspace}

\usepackage{float}
\usepackage{graphicx}
\usepackage{subcaption}

\newcommand{\ours}{\emph{DreamerV3-XP}\xspace}

\title{DreamerV3-XP: Optimizing exploration through uncertainty estimation}

\author{%
  Lukas Bierling\\
  \texttt{lukas.bierling@student.uva.nl} \\
  \And
  Davide Pasero\\
  \texttt{davide.pasero@student.uva.nl} \\
  \And
  Jan Henrik Bertrand\\
  \texttt{jan.henrik.bertrand@student.uva.nl} \\
  \And
  Kiki van Gerwen\\
  \texttt{kiki.van.gerwen@student.uva.nl} \\
}

\begin{document}

\maketitle

\begin{abstract}
We introduce \ours, an extension of DreamerV3 that improves exploration and learning efficiency. This includes (i) a prioritized replay buffer, scoring trajectories by return, reconstruction loss, and value error and (ii) an intrinsic reward based on disagreement over predicted environment rewards from an ensemble of world models. \ours is evaluated on a subset of Atari100k and DeepMind Control Visual Benchmark tasks, confirming the original DreamerV3 results and showing that our extensions lead to faster learning and lower dynamics model loss, particularly in sparse-reward settings. \footnote{All code is available under \hyperlink{https://github.com/Coluding/dreamerv3/}{https://github.com/Coluding/dreamerv3/}}
\end{abstract}

\section*{Introduction}
Recently, several Reinforcement Learning (RL) works investigated using world models to increase sample efficiency and enable planning \cite{hafner2019rssm, hafner2020dreamerv1, hafner2022dreamerv2, hafner2023dreamerv3}. This line of work started with the introduction of a Recurrent State-Space Model (RSSM) \cite{hafner2019rssm} that models environment states as a recurrent latent state for use in predicting future latent states, actions, rewards and values given a policy.

The introduction of the RSSM was a fundamental step in moving away from highly specific RL algorithms that are designed to work well in certain environments (e.g. dense rewards \& continuous actions) while not being applicable to others (e.g. sparse rewards \& discrete actions). In that, Hafner et. al \cite{hafner2023dreamerv3} proposed the DreamerV3, a model-based actor-critic algorithm that is applicable to different domains using one hyperparameter configuration. However, while DreamerV3 demonstrates strong empirical performance, it relies on large-scale models and substantial computational resources. To boost learning and improve sample efficiency, we identify two areas of improvement.
 

First, DreamerV3 samples uniformly from the replay buffer, regardless of a trajectory's value for the learning process. This leaves untapped potential to focus learning on the most informative trajectories, particularly in sparse-reward settings, a well-known drawback in off-policy learning settings~\cite{schaul2015prioritized, kauvar2023curiousreplay}. Inspired by Prioritized Experience Replay ~\cite{schaul2015prioritized} and Curious Replay \cite{kauvar2023curiousreplay}, we hypothesize that a replay scheme incorporating task return, VAE reconstruction error and value error can accelerate learning by prioritizing informative yet uncertain transitions.

Second, exploration is purely guided by environment rewards, implicitly prioritizing trajectories that are already known to be rewarding. Hence, this can hinder broad and thorough exploration, especially in sparse reward settings. To incentivize the exploration of trajectories that are not only promising but also still uncertain in the beginning of the learning process, we introduce an intrinsic reward. It is obtained using the variance among reward predictions from an ensemble of world models as a proxy for uncertainty. Prior work~\cite{sekar2020plan2explore} shows that incorporating epistemic uncertainty via ensemble disagreement improves exploration, but DreamerV3 does not leverage such signals.


We present \ours incorporating a prioritized replay buffer and an uncertainty-driven exploration reward to improve exploration and data efficiency.

\section*{Background}

\paragraph{Related Work }
In the realm of Dreamer-based RL, advances in exploration and replay buffer design have been central. For \emph{exploration}, Plan2Explore~\cite{sekar2020plan2explore} introduces intrinsic rewards by computing disagreement among an ensemble of world models, enabling unsupervised skill acquisition and robust zero-shot transfer. PEG~\cite{hu2023planninggoalsexploration} further improves exploration by selecting and planning toward goal states with high novelty and epistemic uncertainty. In parallel, model-free methods based on curiosity-driven objectives~\cite{pathak2017curiositydrivenexplorationselfsupervisedprediction} promote exploration through prediction error, a principle now widely adopted.
Regarding \emph{replay buffer optimization}, Curious Replay~\cite{kauvar2023curiousreplay} prioritizes trajectories with high model prediction error, thereby improving sample efficiency in sparse-reward settings, while Prioritized Experience Replay (PER)~\cite{schaul2015prioritized} remains influential for adaptive sampling across RL algorithms.
These directions collectively underscore the importance of intrinsic motivation and informed replay in efficient model-based RL.

\paragraph{DreamerV3}
At the core of DreamerV3’s architecture lies a world model that enables planning via imagination. This model is implemented as a Recurrent State-Space Model (RSSM) \cite{hafner2019rssm}, which learns compact latent dynamics by combining stochastic and deterministic components.
Specifically, the model encodes observations using a variational autoencoder \cite{kingma2013auto} to obtain a stochastic latent state $z_t$. This latent, together with the previous action $a_{t-1}$ and the recurrent hidden state $h_{t-1}$, is used to update the deterministic state via $h_t = f_\phi(h_{t-1}, z_{t-1}, a_{t-1})$.
The pair $(h_t, z_t)$ forms the full latent state used for prediction. Given this representation, the model can generate future trajectories without explicit environment interaction, predicting rewards, continuation probabilities, and observations conditioned solely on hypothetical actions. This process is referred to as imagination.
This enables Dreamer’s actor–critic to be trained entirely in latent space, i.e. in imagination, significantly improving sample efficiency and enabling long-horizon foresight.

\section*{Methods}

\paragraph{Reproduction Study}
DreamerV3 is evaluated across a broad benchmark suite spanning over 150 tasks from 8 domains, including both discrete and continuous action spaces, as well as dense and sparse reward settings. Due to computational constraints, we focus our reproduction efforts on two representative benchmarks: Atari100k ~\cite{bellemare2013arcade} and the DeepMind Control Vision Benchmark ~\cite{ortiz2024dmcvbbenchmark}. These were selected to cover both discrete-action, dense reward settings (Atari100k) and continuous-action, sparse reward tasks (DMC Vision). All experiments are run with two random seeds.

From Atari100k, we selected \emph{Battle Zone, Boxing, and Krull}, to reflect diversity in reward structure and exploration difficulty. Boxing requires a very precise world model to estimate the position relative to the opponent. Krull covers several different landscapes and thus requires a complex but efficient world model which models relevant parts of the environment precisely. Battle Zone provides dense rewards for shooting enemies but requires sustained planning to survive escalating waves.

From DMC Vision, we chose \emph{Reacher (hard) and Cup Catch}, two sparse reward tasks that differ in difficulty and stability. Cup Catch comes with a vast space of possible trajectories of which only a small subset will land the ball into the cup, making it ideal for testing prioritized exploration. Reacher (hard) requires accurately steering a two-joint arm to a precise location which yields the reward.

\paragraph{Latent Reward Disagreement}

By default, DreamerV3 explores using its policy trained on environment rewards (which we will refer to as "extrinsic rewards") without taking uncertainty estimates into account. To incentivize the exploration of uncertain but promising trajectories, we mix in a so-called intrinsic reward in order to allow for more exploration while still prioritizing relevant trajectories.


Inspired by Plan2Explore's \cite{sekar2020plan2explore} "disagreement" over latent states predicted by an ensemble of world models, we use the disagreement over reward predictions from an ensemble of world models. To quantify the disagreement, the variance over the predicted rewards is taken and added to the mean of the predicted rewards to incentivize trajectories that are expected to be rewarding. This sum of mean and variance is our intrinsic reward. 
Each ensemble member $k \in \{1, .., K\}$, parameterized by $w_k$, recurrently predicts (i.e., "imagines") future deterministic latent states $h_{t'}^{w_k}$ over imagination horizon $L$ with $t'$ being a timestep within the horizon. The standard reward predictor then predicts the corresponding reward $\hat{r}_{k, t'} \sim p_\phi(\hat{r}_{t'} | h_{t'}^{w_k}, z_{t'})$. Formally,
$$
r_{t}^{intr} = \frac{1}{L} \sum_{t'=t}^{t+L} \left[ \bar{r}_{t'} + \frac{1}{K} \sum_{k=1}^{K} (\hat{r}_{k,t'} - \bar{r}_{t'})^2 \right]
$$
where $\bar{r}_{t'}$ is the mean predicted reward across all ensemble members at timestep $t'$ of the imagination. High variance indicates epistemic uncertainty over the predicted reward, and thus encourages exploration of the associated state. The final reward used for training is a convex combination of extrinsic and intrinsic rewards: 
$$r_t^{\text{total}} = \lambda r_t^{\text{ext}} + (1-\lambda) r_t^{\text{intr}}$$

\paragraph{Dynamic Reward Weighting}  
To balance exploration and exploitation, we combine extrinsic and intrinsic rewards using a weighting factor \(\lambda\). We experiment with two strategies for adapting \(\lambda\) over time. First, we apply exponential decay, gradually reducing the influence of intrinsic rewards as training progresses. Second, we explore a dynamic adjustment using the gradient of an exponential moving average (EMA) of the episode return: \(\lambda\) is decreased when performance tends to improve and increased when learning stagnates or regresses. This encourages exploration when necessary and promotes exploitation when training is stable.

\paragraph{Optimized Replay}

To improve learning efficiency, we introduce a prioritized sampling strategy that scores trajectories using a weighted combination of task relevance, VAE reconstruction error, and critic value error. The priority score \(s_i\) for each trajectory \(i\) is defined as:
\[
s_i = (\lambda_r + \lambda_\delta \delta_i) R_i + \lambda_\epsilon \epsilon_i,
\]
where \(R_i\) is the total return, \(\epsilon_i\) the VAE reconstruction error, and \(\delta_i\) the critic value error. This approach focuses updates on rewarding transitions where the agent's predictions are least certain and seem
to be valuable for the task.

\section*{Results}
\paragraph{Reproduction}

\begin{figure}[H]
    \centering
    \begin{subfigure}[t]{0.195\textwidth}
        \includegraphics[width=\linewidth]{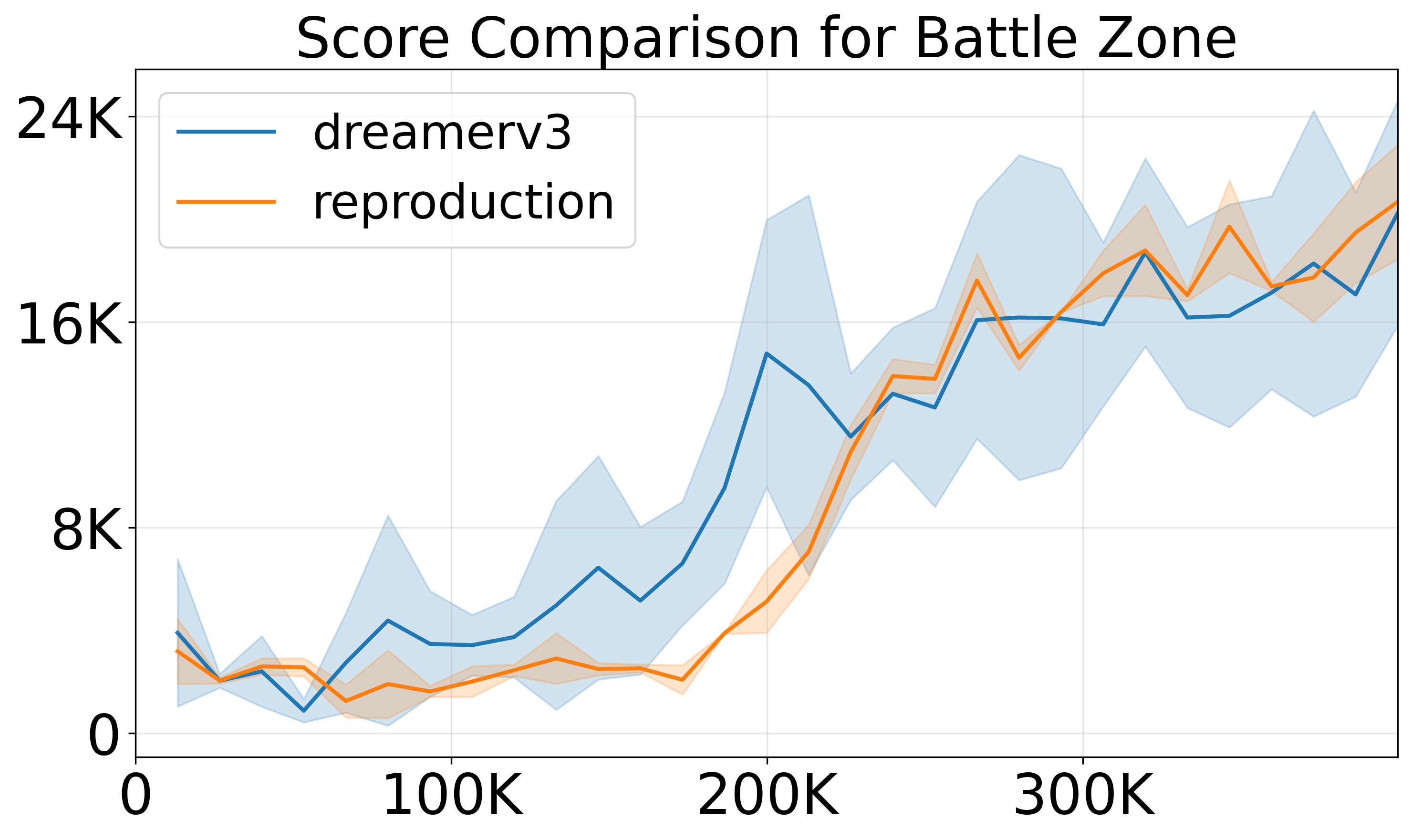}
        \caption{Battle Zone}
    \end{subfigure}
    \begin{subfigure}[t]{0.195\textwidth}
        \includegraphics[width=\linewidth]{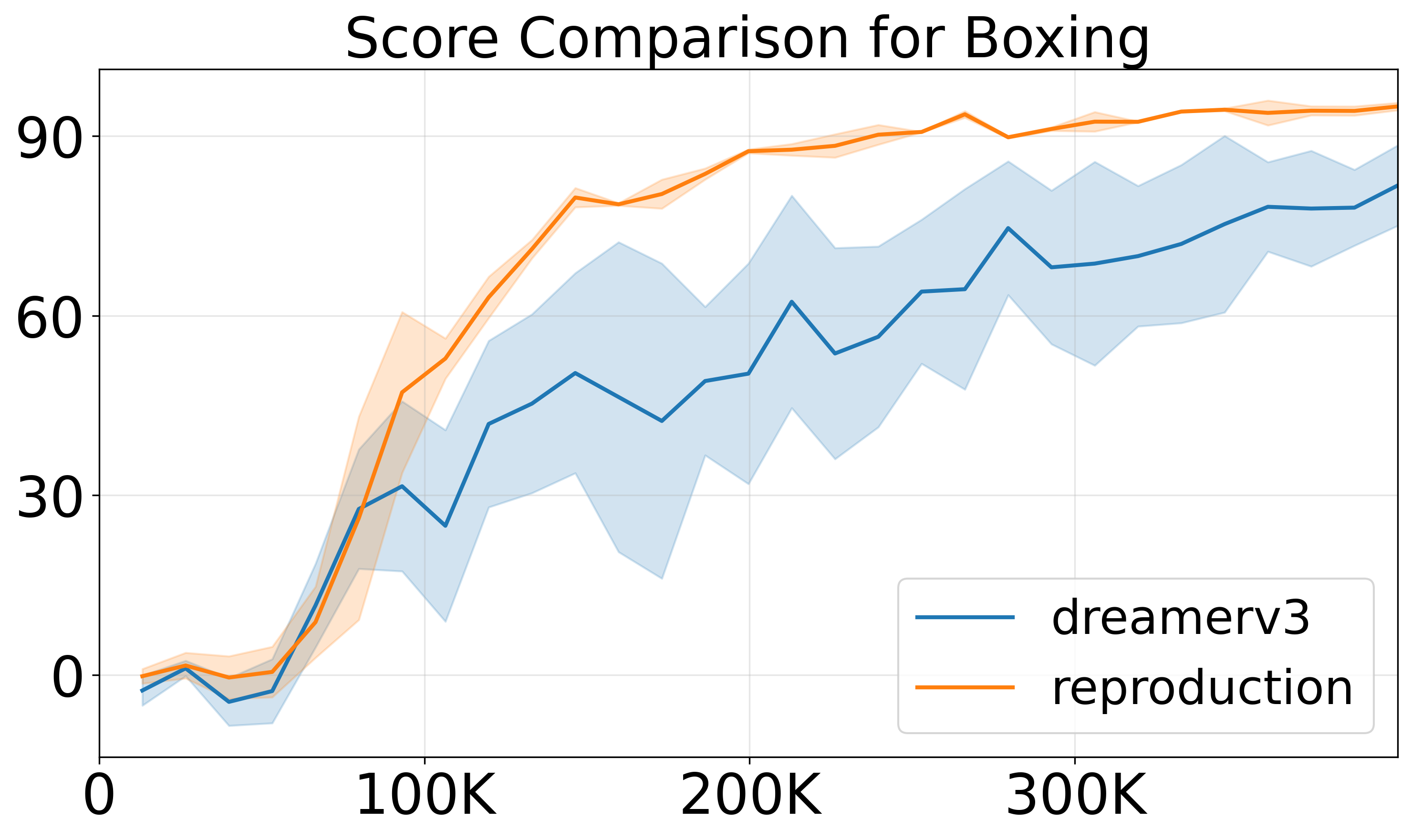}
        \caption{Boxing}
    \end{subfigure}
    \begin{subfigure}[t]{0.195\textwidth}
        \includegraphics[width=\linewidth]{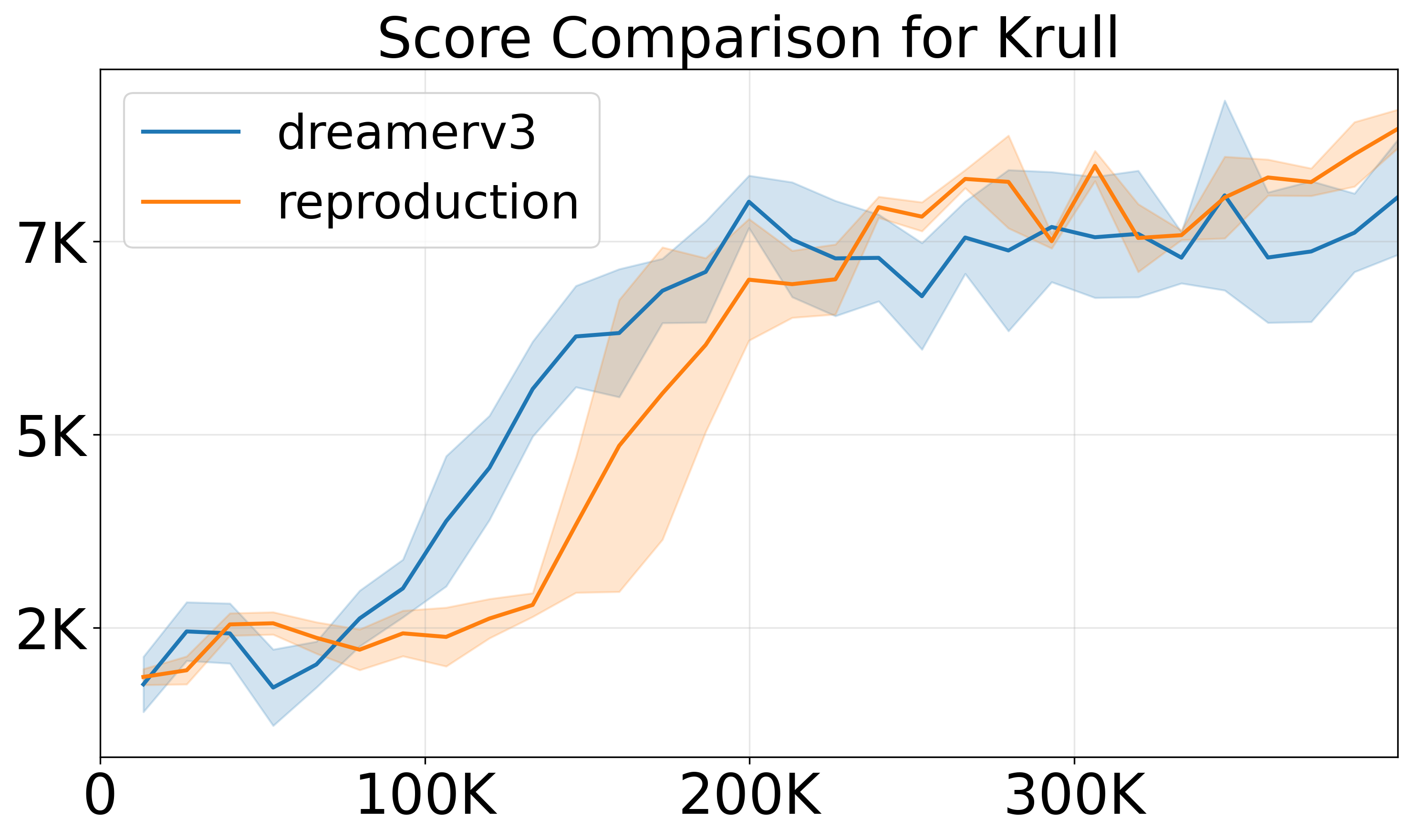}
        \caption{Krull}
    \end{subfigure}
    \begin{subfigure}[t]{0.195\textwidth}
        \includegraphics[width=\linewidth]{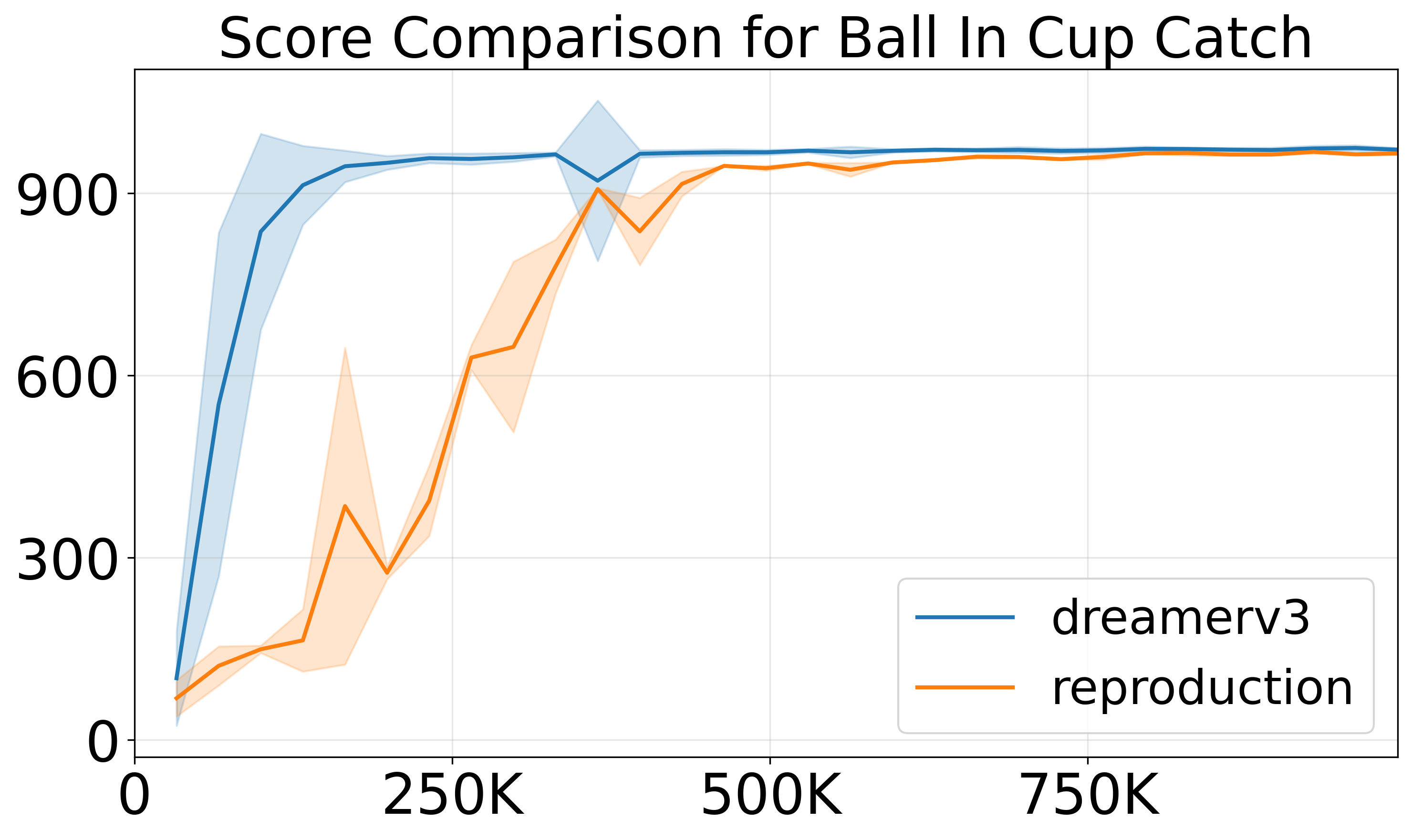}
        \caption{Cup Catch}
    \end{subfigure}
    \begin{subfigure}[t]{0.195\textwidth}
        \includegraphics[width=\linewidth]{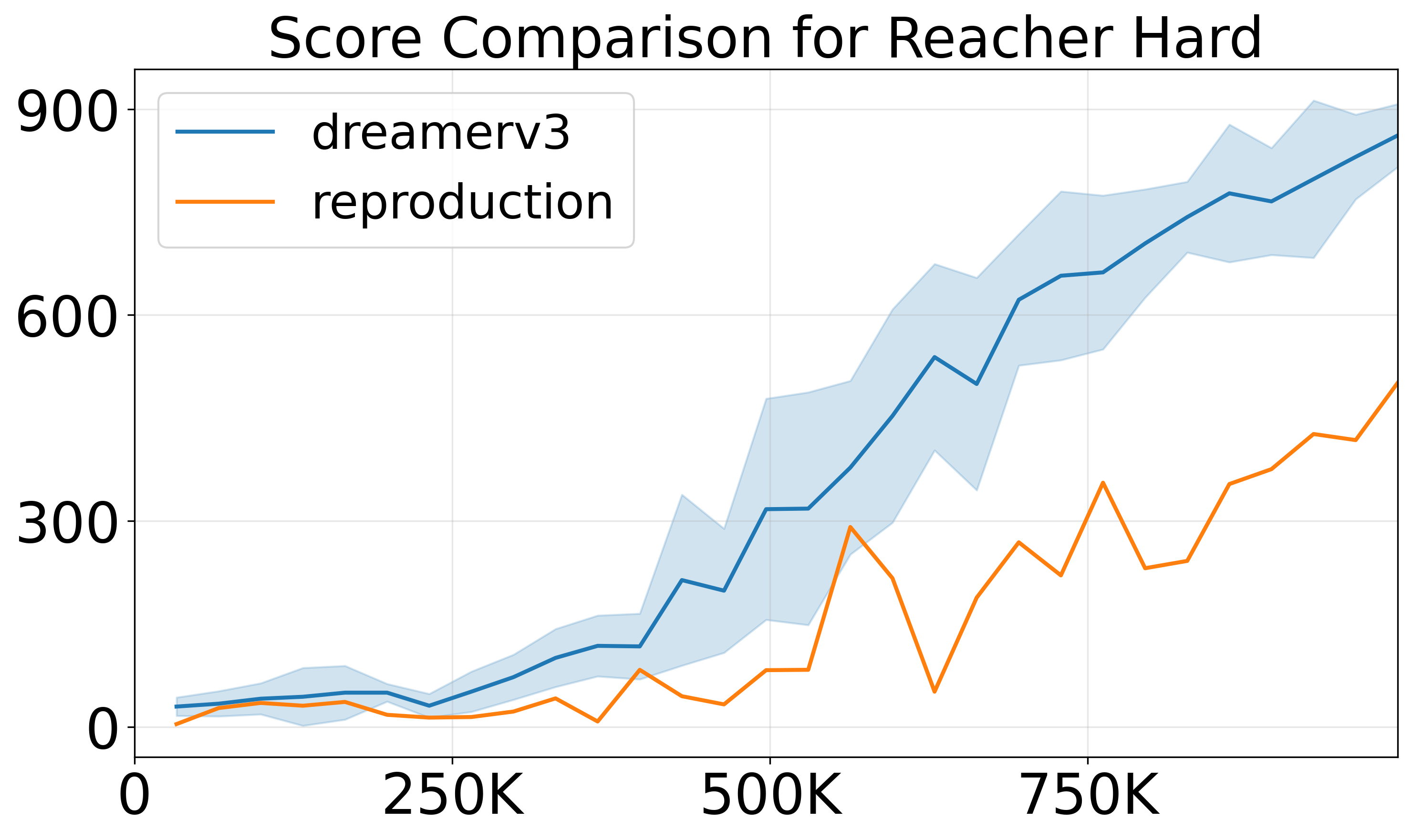}
        \caption{Reacher (hard)\textasteriskcentered}
    \end{subfigure}
    \caption{Reproduction results across five tasks. Mean episode return over 2 seeds is shown (with shaded variance), except for \textasteriskcentered, which uses a single seed due to computational constraints.}
    \label{fig:repro_comparison}
\end{figure}

Table~\ref{fig:repro_comparison} compares our reproduced episode scores with those reported by the DreamerV3 authors across five selected tasks. Some of our results deviate which however can be attributed to variance from fewer runs (we used two seeds compared to five seeds in the DreamerV3 paper). In summary, we can mostly confirm DreamerV3’s reported performance.

\vspace{-1em}
\vspace{-1.5em}

\begin{figure}[H]
    \centering
    \begin{subfigure}[t]{0.25\textwidth}
        \centering
        \includegraphics[width=\linewidth]{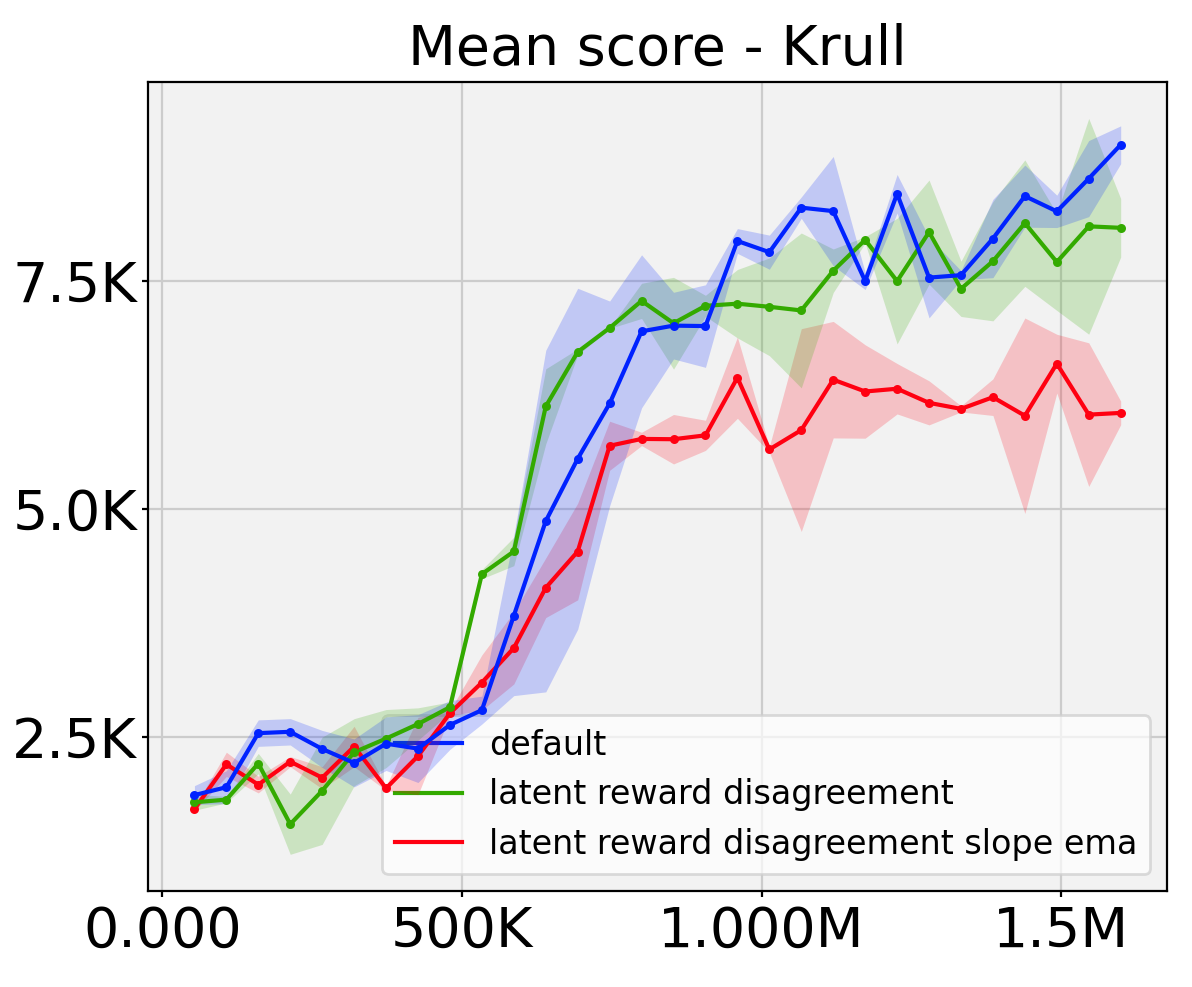}
        \caption{Krull episode scores}
        \label{fig:lrd_krull}
    \end{subfigure}
    \hspace{2em}
    \begin{subfigure}[t]{0.25\textwidth}
        \centering
        \includegraphics[width=\linewidth]{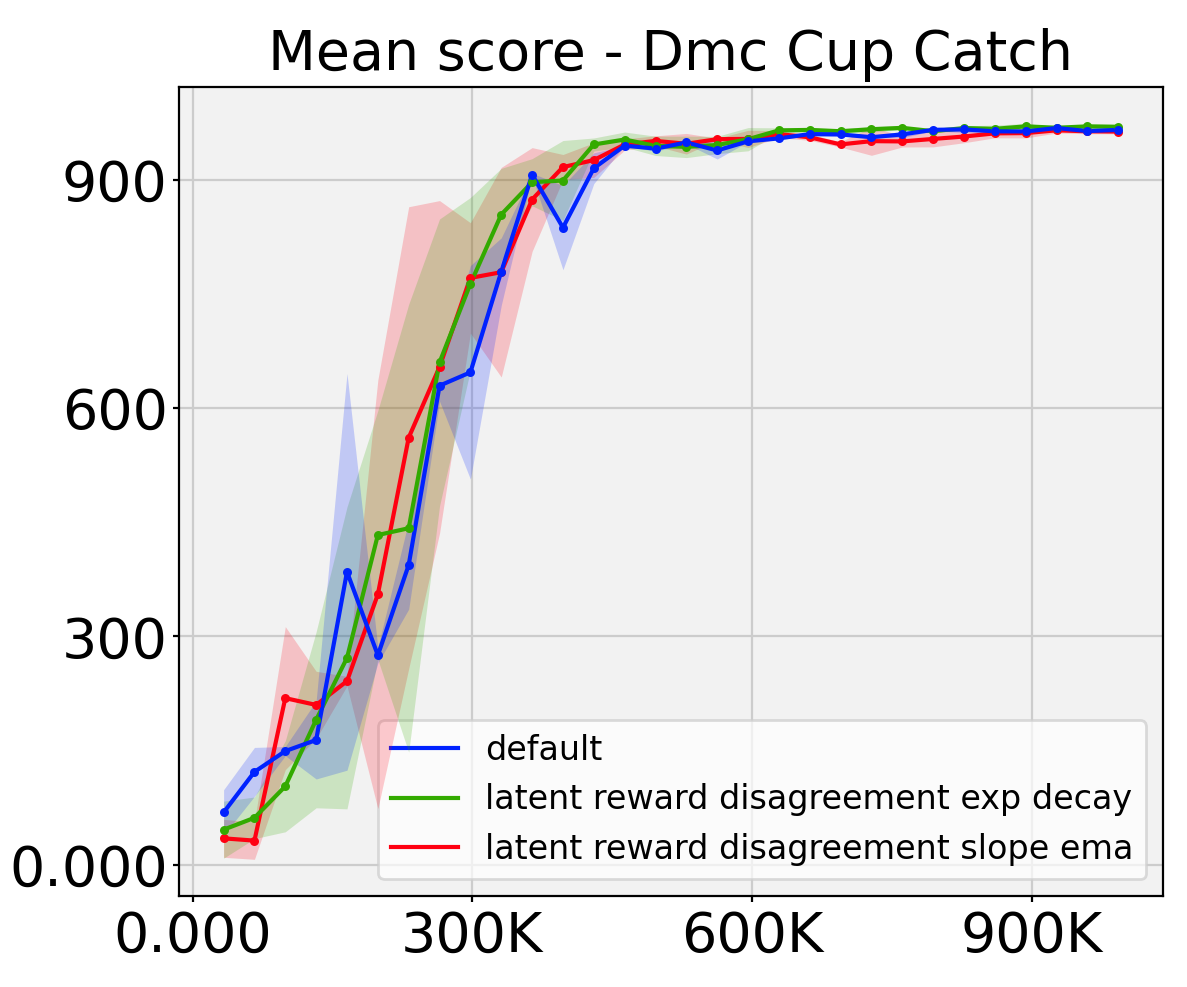}
        \caption{Cup Catch episode score}
        \label{fig:lrd_cup_catch}
    \end{subfigure}
    \caption{Mean (2 seeds) episode scores for Krull and Cup Catch with reward disagreement.}
    \label{fig:lrd_performance}
\end{figure}

\paragraph{Latent Reward Disagreement}
Figure ~\ref{fig:lrd_performance} suggests that our basic latent reward disagreement mechanism, with a constant $\lambda$, leads to slight improvements in learning speed on Krull, with slightly steeper episode return curves early in training. Applying an exponential decay to $\lambda$, or scaling it based on the slope of the EMA, yields mildly positive effects for Cup Catch. Notably, one of the EMA runs for Cup Catch exhibited a much steeper learning curve, indicating potential benefits worth exploring further.\footnote{Due to resource constraints, we were unfortunately not able to extend this to more seeds and tasks.} While the observed gains are modest, they are consistent with the hypothesis that guiding exploration toward uncertain but potentially reward-relevant states can facilitate progress.

\vspace{-1em}
\vspace{-1.5em}

\begin{figure}[H]
    \centering
    \begin{subfigure}[t]{0.195\textwidth}
        \includegraphics[width=\linewidth]{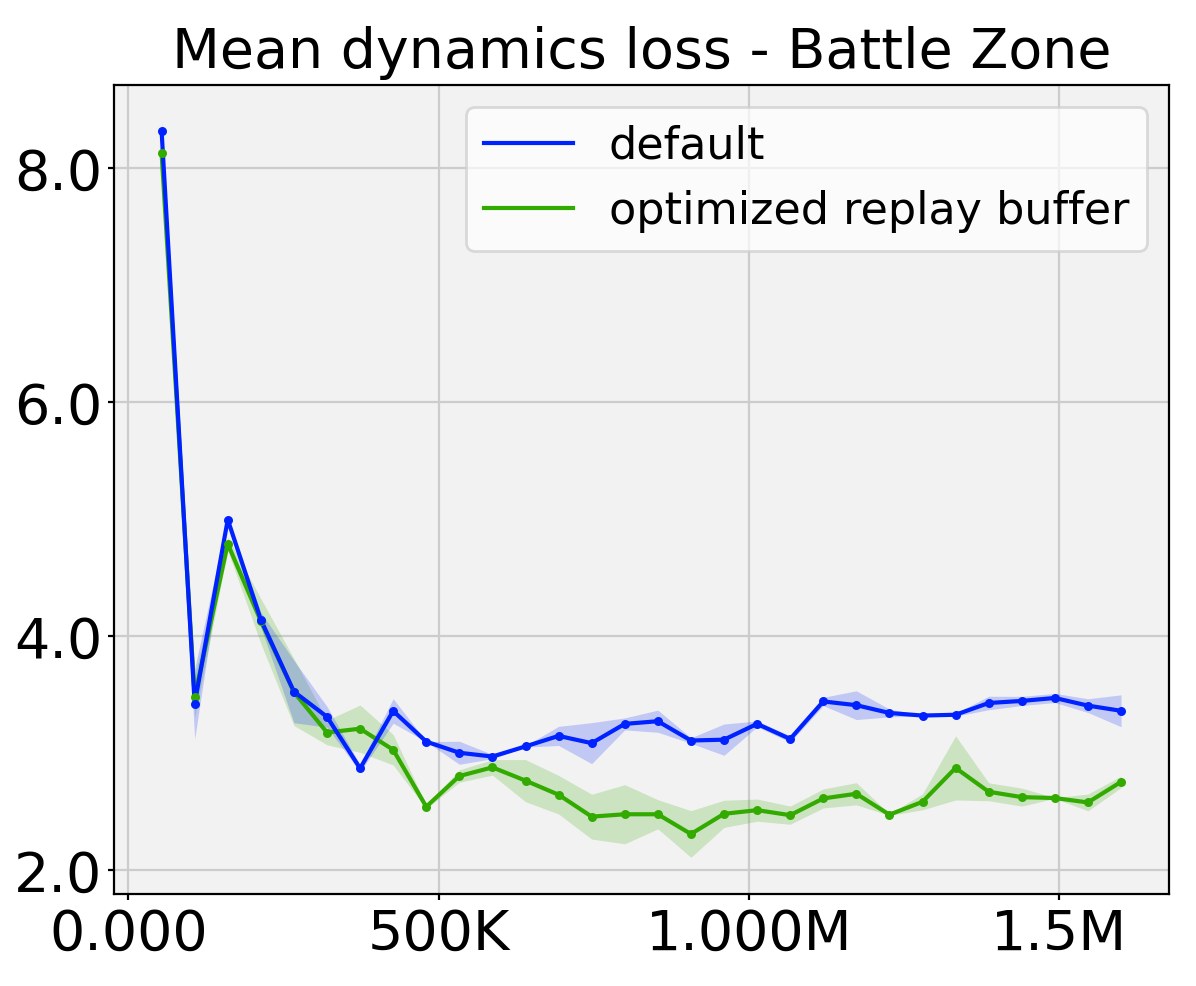}
        \caption{Battle Zone}
    \end{subfigure}
    \begin{subfigure}[t]{0.195\textwidth}
        \includegraphics[width=\linewidth]{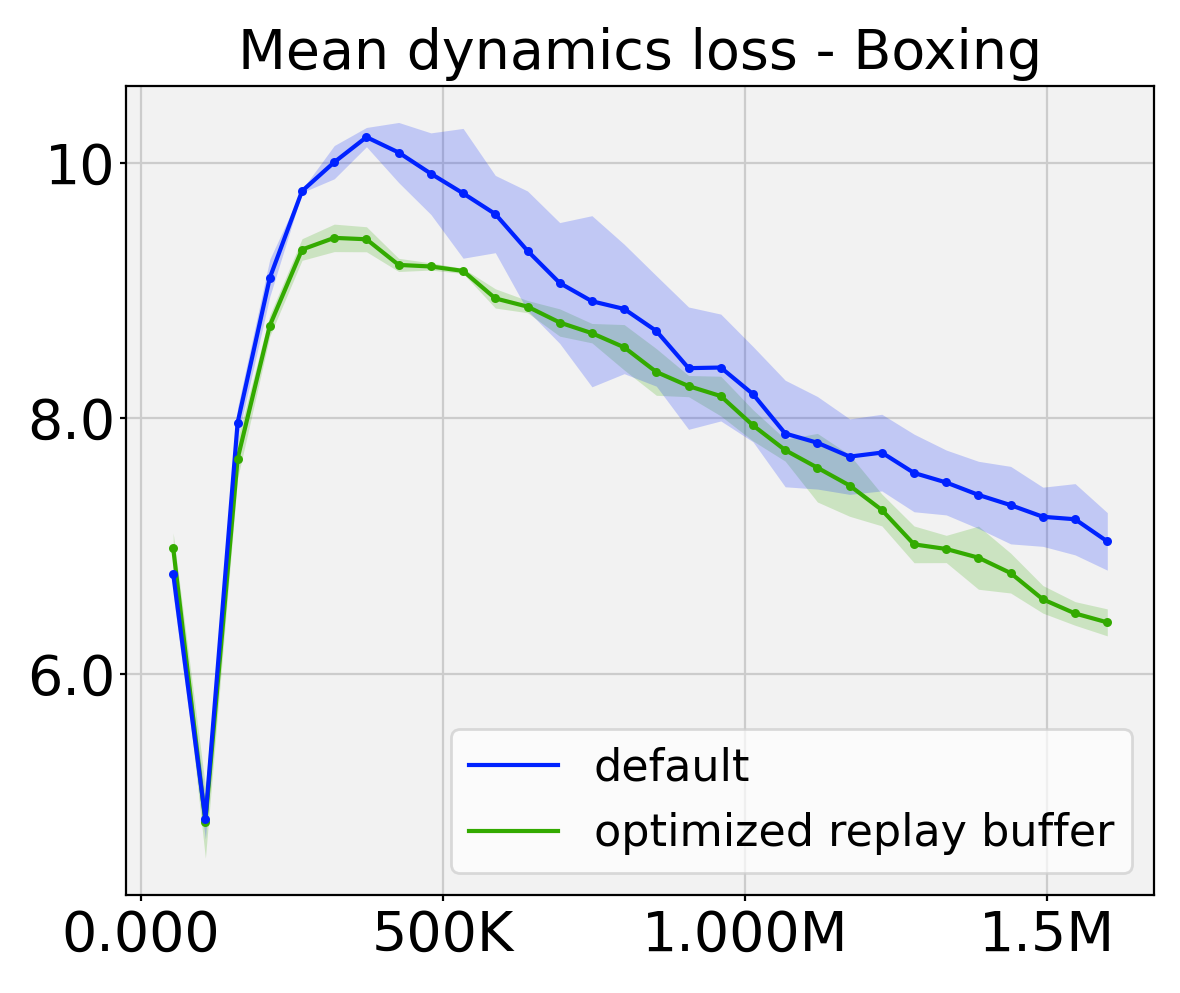}
        \caption{Boxing}
    \end{subfigure}
    \begin{subfigure}[t]{0.195\textwidth}
        \includegraphics[width=\linewidth]{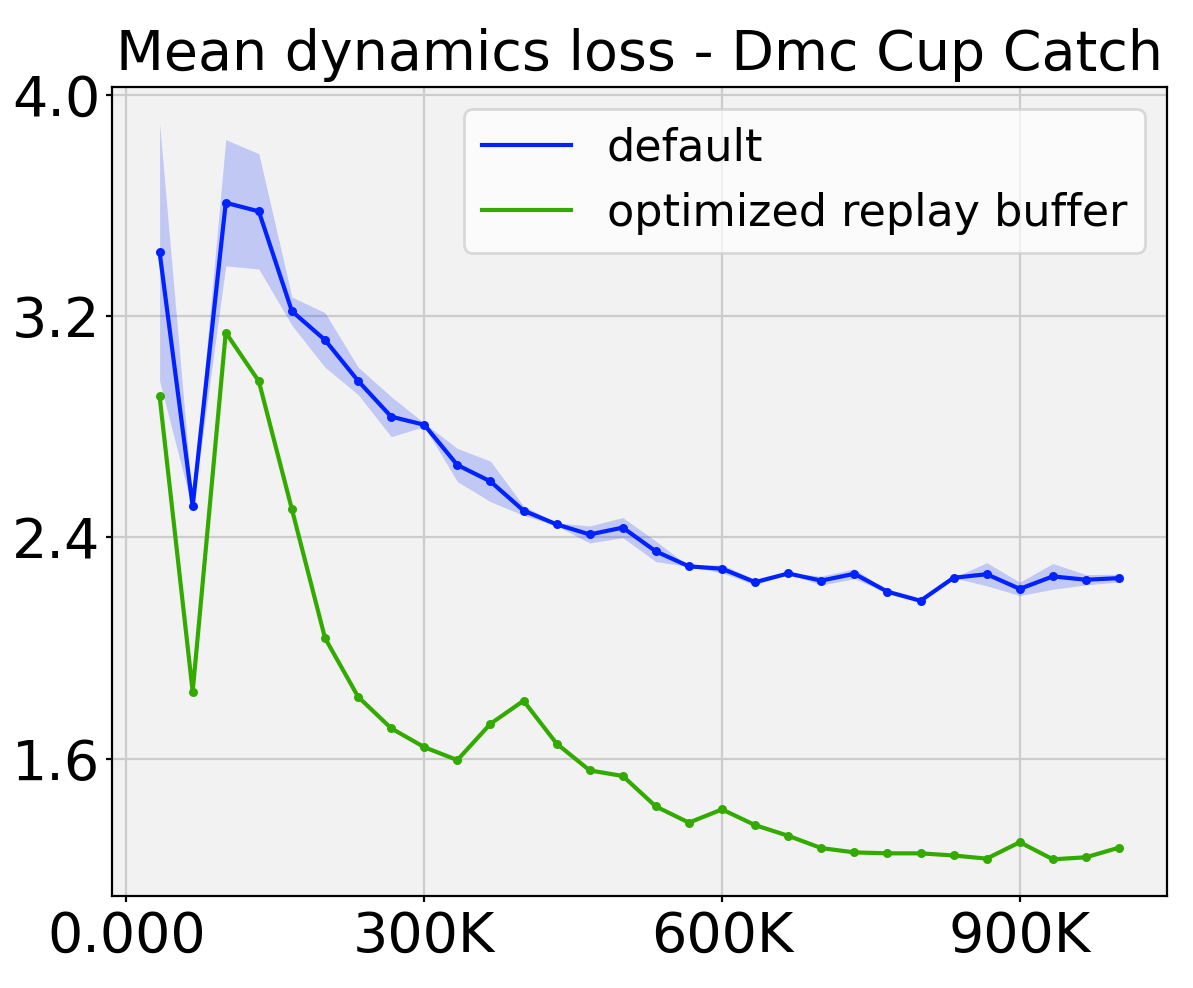}
        \caption{Cup Catch \textasteriskcentered}
    \end{subfigure}
    \begin{subfigure}[t]{0.195\textwidth}
        \includegraphics[width=\linewidth]{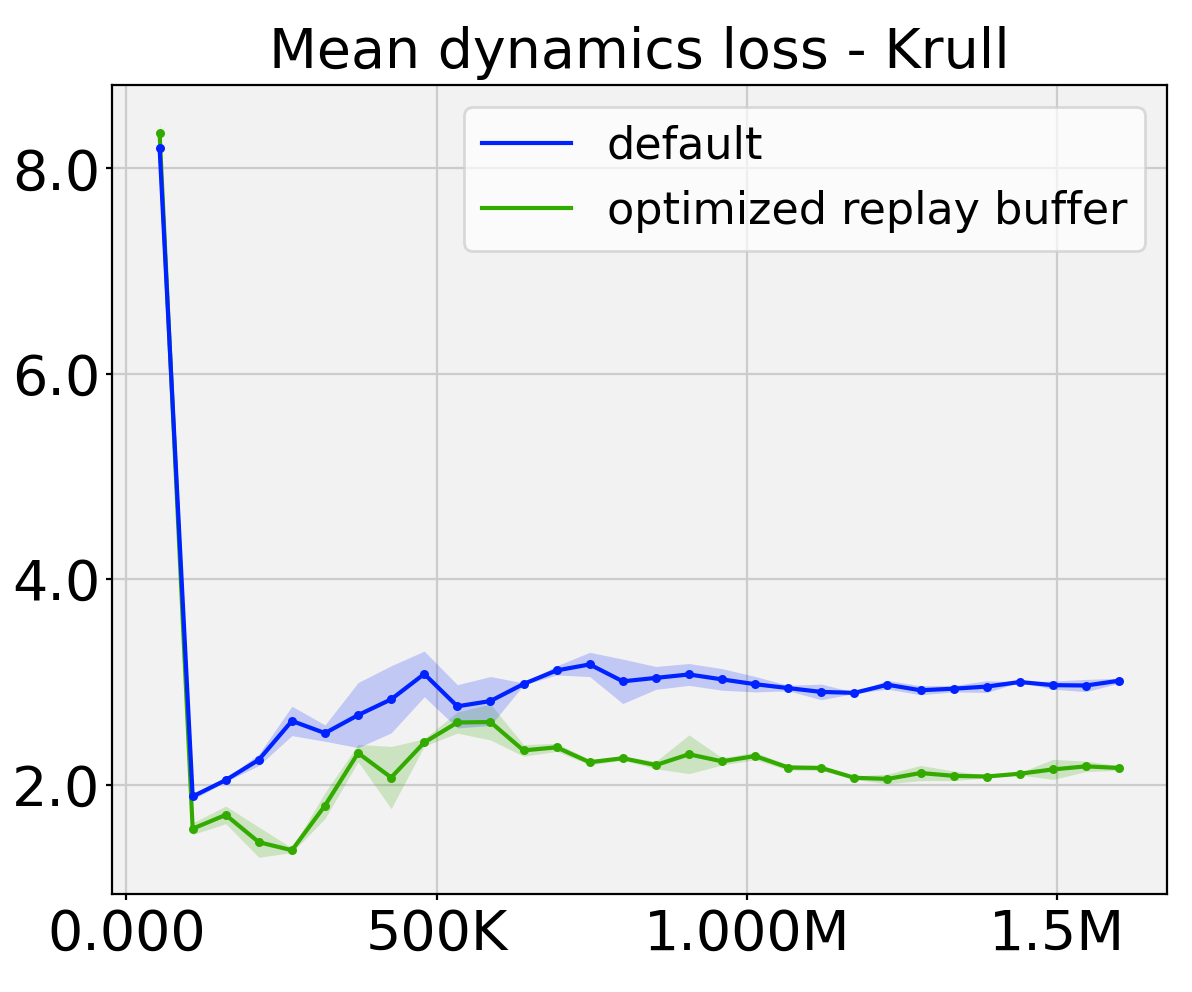}
        \caption{Krull}
    \end{subfigure}
    \begin{subfigure}[t]{0.195\textwidth}
        \includegraphics[width=\linewidth]{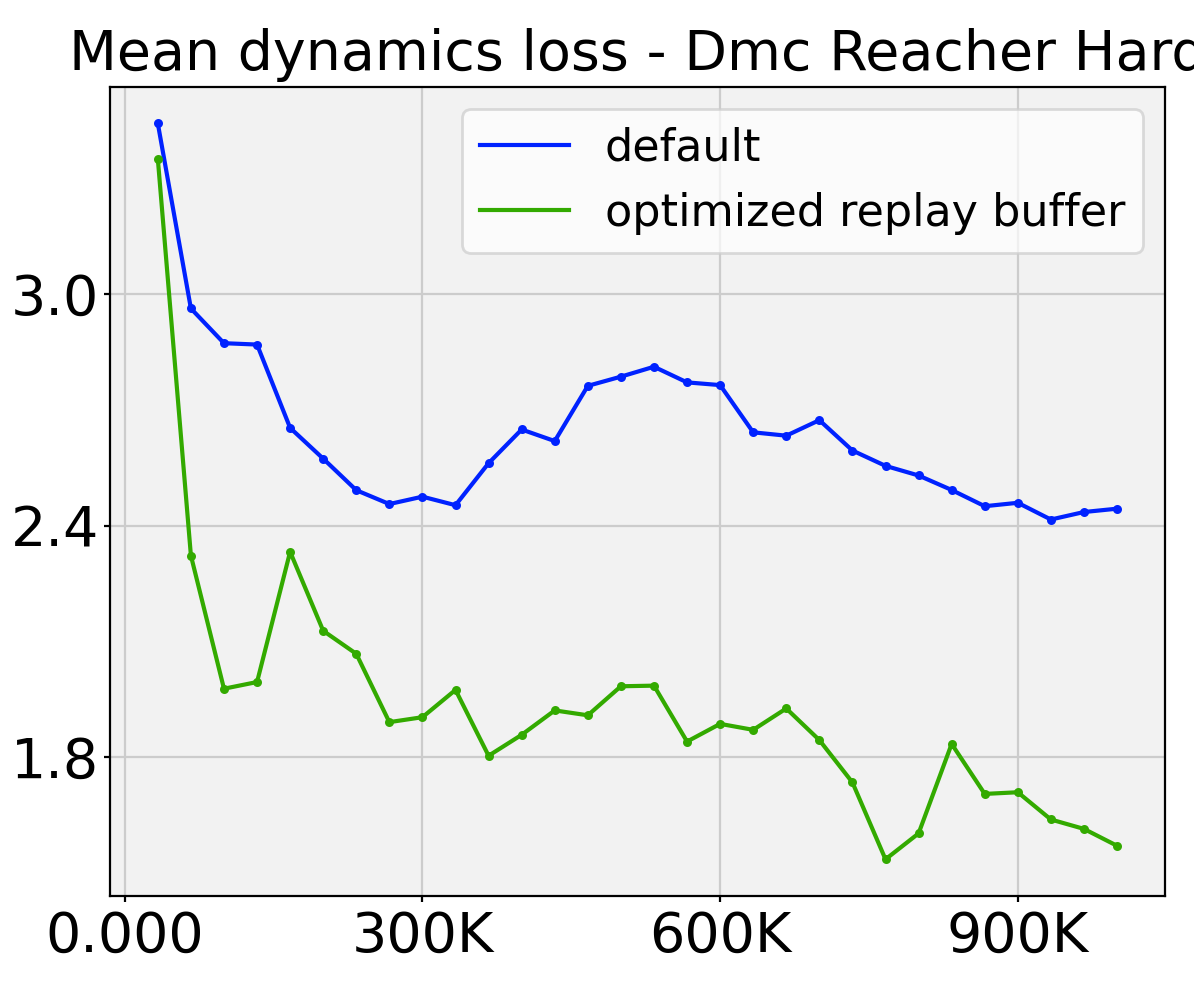}
        \caption{Reacher (hard) \textasteriskcentered}
    \end{subfigure}
    \caption{Mean dynamics model training loss across five tasks, comparing optimized replay to the default setting. Results are averaged over 2 seeds, except for \textasteriskcentered, which uses a single seed due to computational constraints.}
    \label{fig:dynamics_losses}
\end{figure}

\paragraph{Optimized Replay}
Figure~\ref{fig:dynamics_losses} shows that our optimized replay consistently reduces the dynamics loss across all tested tasks. This suggests that the world model learns more accurate latent transitions, resulting in more reliable imagined rollouts and policy learning.

\begin{figure}[h!]
    \centering
    \begin{subfigure}[t]{0.22\textwidth}
        \centering
        \includegraphics[width=\linewidth]{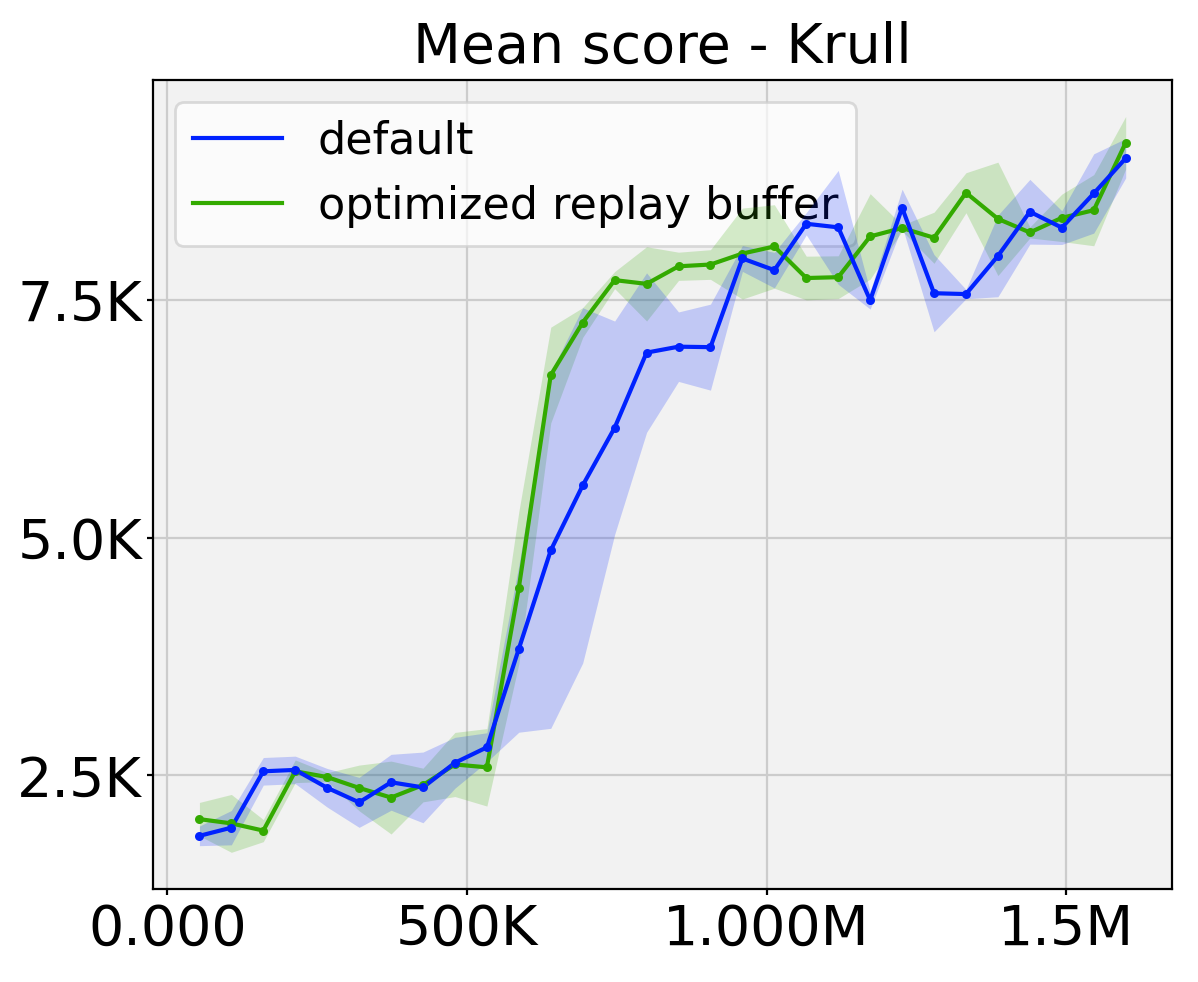}
        \caption{Episode return}
        \label{fig:krull_perf}
    \end{subfigure}
    \hspace{1em}
    \begin{subfigure}[t]{0.22\textwidth}
        \centering
        \includegraphics[width=\linewidth]{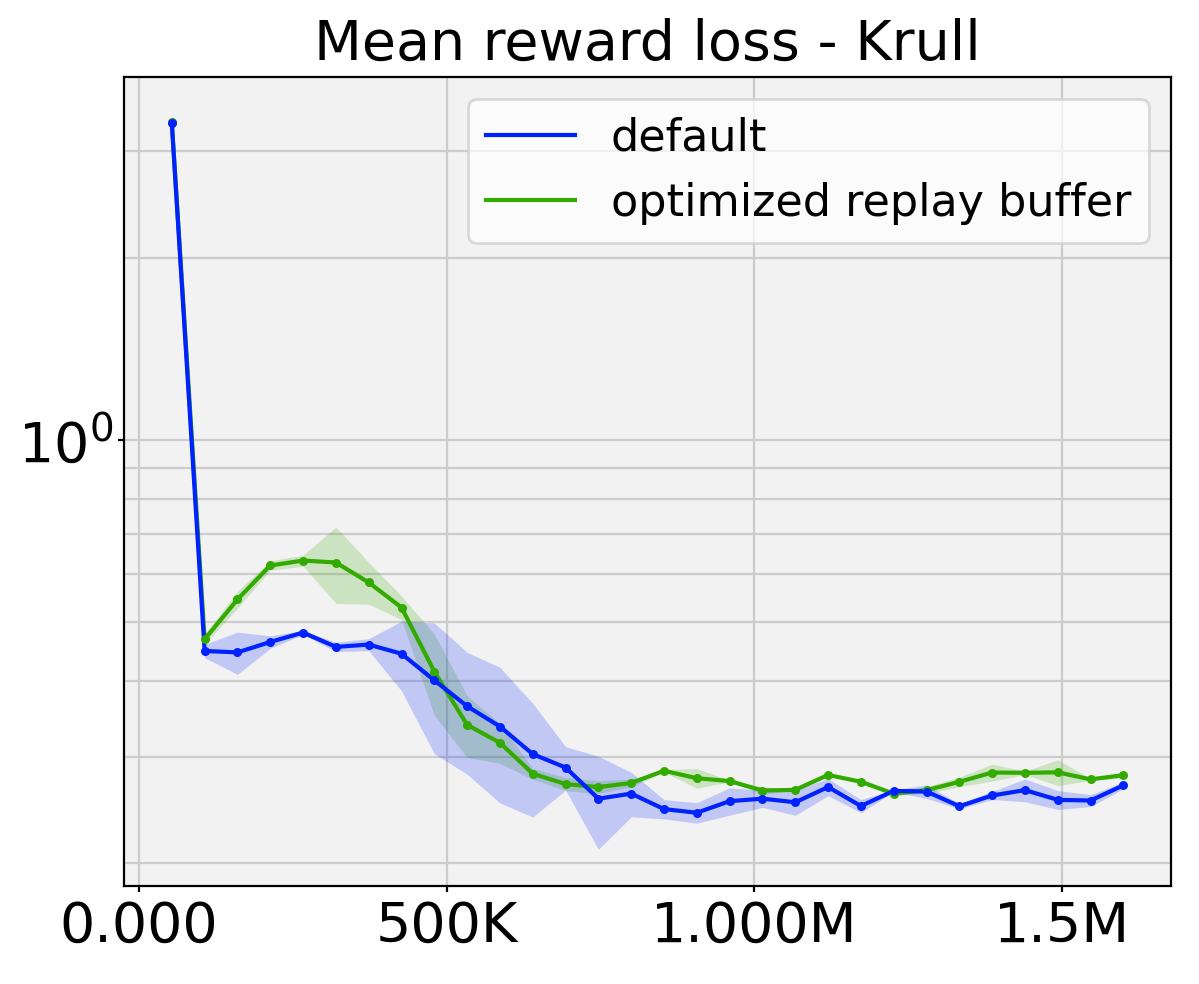}
        \caption{Reward loss}
        \label{fig:krull_rew_loss}
    \end{subfigure}
    \hspace{1em}
    \begin{subfigure}[t]{0.22\textwidth}
        \centering
        \includegraphics[width=\linewidth]{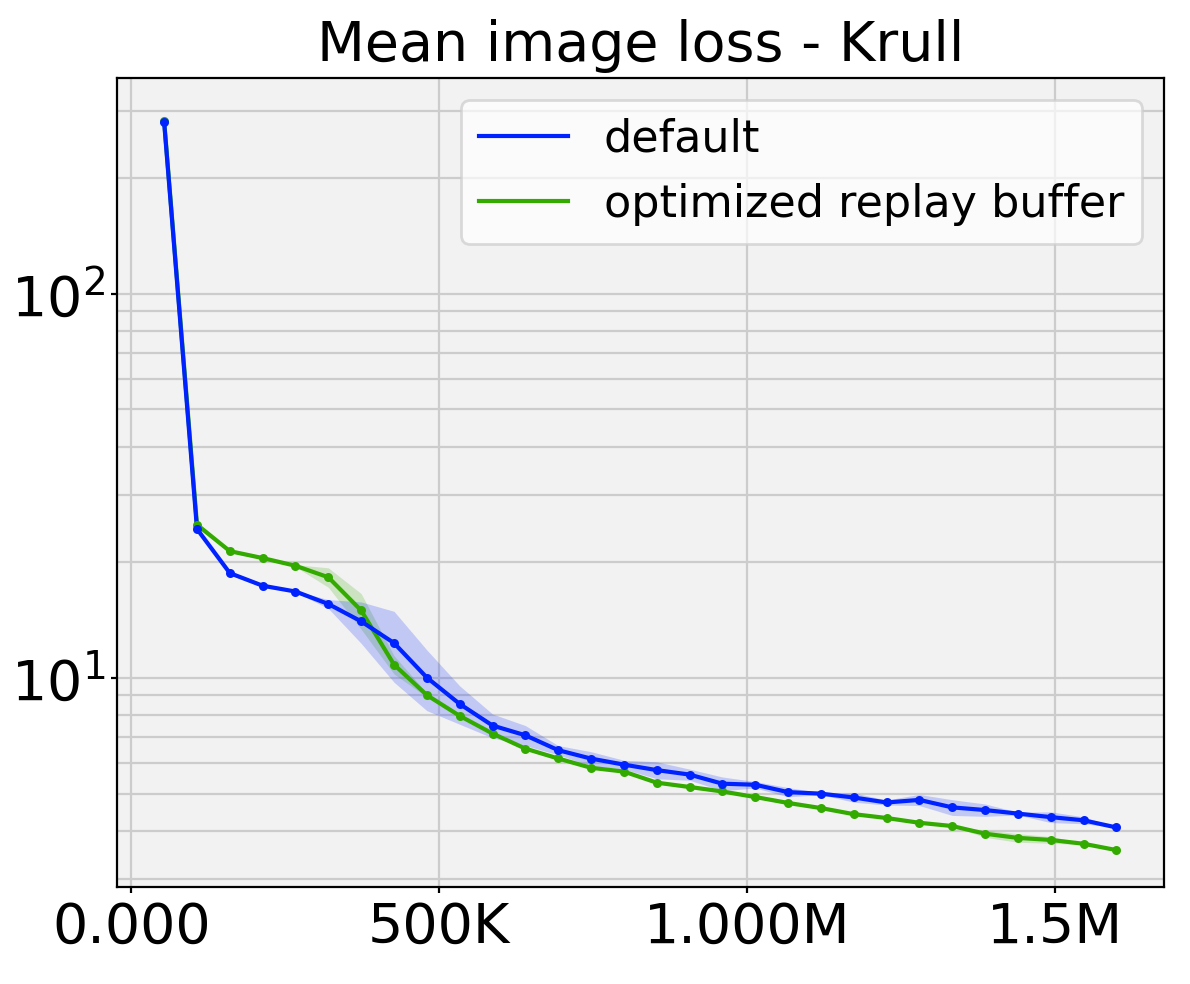}
        \caption{Reconstruction loss}
        \label{fig:krull_img_loss}
    \end{subfigure}
    \hspace{1em}
    \begin{subfigure}[t]{0.22\textwidth}
        \centering
        \includegraphics[width=\linewidth]{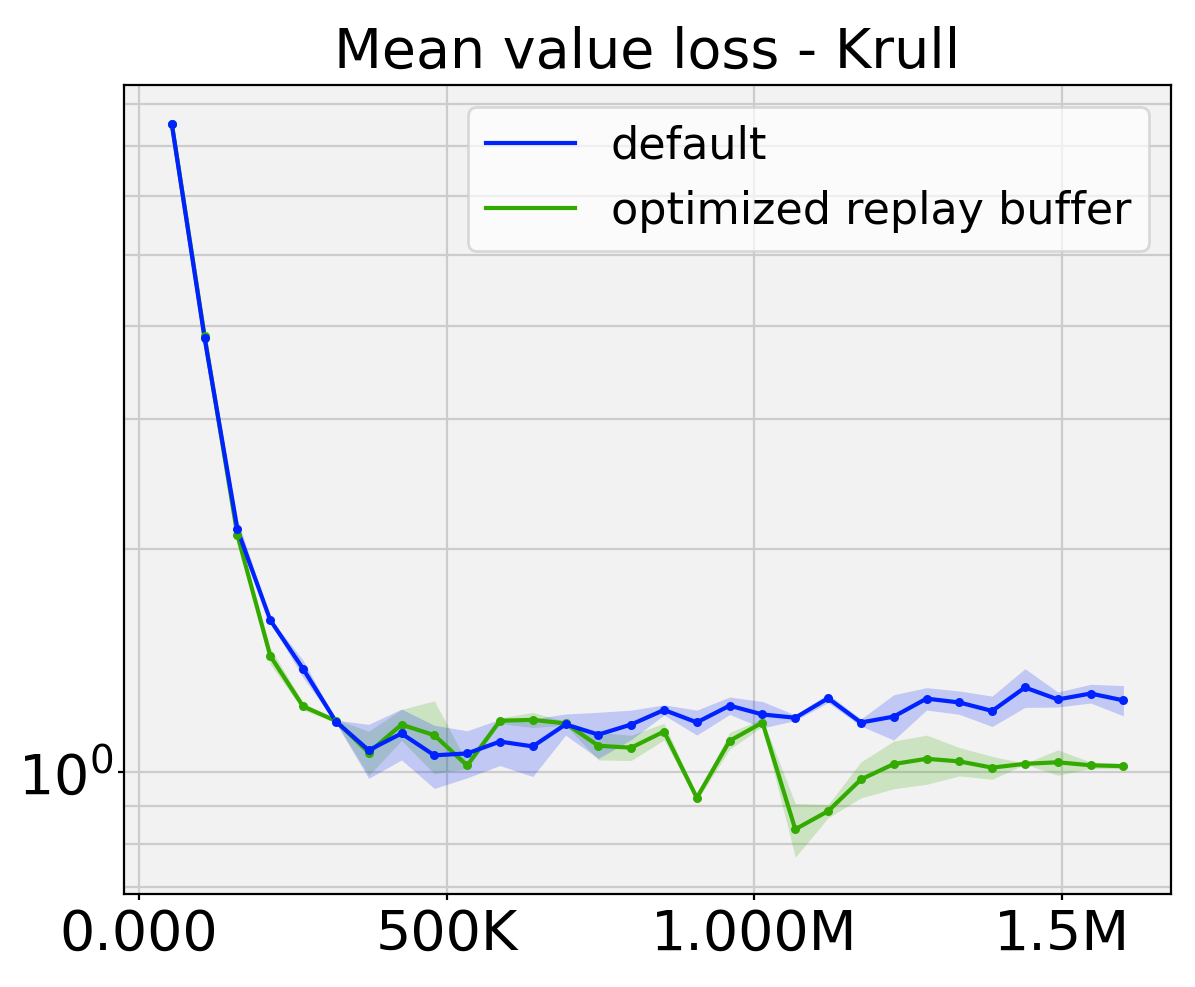}
        \caption{Value loss}
        \label{fig:krull_val_loss}
    \end{subfigure}

    \caption{Mean (2 seeds) performance and loss metrics on Krull for optimized replay and default.}
    \label{fig:krull_results}
\end{figure}

In Figure~\ref{fig:krull_results}, we observe lower reconstruction, value, and reward losses when using optimized replay. These results support the intended effect of our prioritization strategy: by sampling trajectories with high reconstruction and value error, the agent focuses updates on underrepresented and challenging transitions. This leads to improved model learning, as evident by stronger latent representations and more accurate predictions. Additionally, the episode return curve rises more steeply early in training, indicating faster learning when updates are focused on more informative experiences.

These results align with our initial hypothesis that uniform sampling dilutes important learning signals, and that prioritizing trajectories based on reconstruction error, value error, and return would help the agent focus on parts of the experience buffer that are both informative and underexplored. The consistent improvements across loss metrics and learning speed support the effectiveness of this targeted sampling strategy.

\vspace{-1em}
\vspace{-1.5em}

\vspace{-1.5em}

\section*{Conclusion}
We presented \ours, extending DreamerV3 with prioritized replay and a latent reward disagreement. Prioritized replay consistently reduced dynamics loss, confirming our hypothesis that prioritizing trajectories based on reconstruction error, value error, and return improves model learning by helping the agent focus on parts of the experience buffer that are both informative and underexplored. While the gains from latent reward disagreement were modest, they suggest potential for further research, such as evaluating its impact across a broader range of tasks and environments. Our findings still support the hypothesis that leveraging epistemic uncertainty can aid exploration and contribute to learning a more effective world model.

\newpage
\bibliographystyle{plain}  
\bibliography{references}

\end{document}